\documentclass{bmvc2k}

\usepackage{multirow}
\usepackage{amssymb}
\usepackage{textcomp}
\usepackage{amsmath}
\usepackage{hyperref}
\DeclareMathOperator*{\argmax}{arg\,max}

\newcommand{\bm}{\mathbf}

\title{Embodied Vision-and-Language Navigation\\with Dynamic Convolutional Filters}

\addauthor{Federico Landi}{federico.landi@unimore.it}{1}
\addauthor{Lorenzo Baraldi}{lorenzo.baraldi@unimore.it}{1}
\addauthor{Massimiliano Corsini}{massimiliano.corsini@unimore.it}{1}
\addauthor{Rita Cucchiara}{rita.cucchiara@unimore.it}{1}

\addinstitution{
 University of Modena and\\Reggio Emilia,
 Italy
}

\runninghead{Landi, et al}{Embodied VLN with Dynamic Convolutional Filters}

\def\eg{\emph{e.g}\bmvaOneDot}

\def\ie{\emph{i.e}\bmvaOneDot}
\def\etal{\emph{et al}\bmvaOneDot}

\begin{document}
\maketitle
%-------------------------------------------------------------------------
\vspace{-.15cm}
\begin{abstract}
In Vision-and-Language Navigation (VLN), an embodied agent needs to reach a target destination with the only guidance of a natural language instruction. To explore the environment and progress towards the target location, the agent must perform a series of low-level actions, such as rotate, before stepping ahead.
In this paper, we propose to exploit dynamic convolutional filters to encode the visual information and the lingual description in an efficient way.
Differently from some previous works that abstract from the agent perspective and use high-level navigation spaces, we design a policy which decodes the information provided by dynamic convolution into a series of low-level, agent friendly actions.
Results show that our model exploiting dynamic filters performs better than other architectures with traditional convolution, being the new state of the art for embodied VLN in the low-level action space.
Additionally, we attempt to categorize recent work on VLN depending on their architectural choices and distinguish two main groups: we call them \textit{low-level actions} and \textit{high-level actions} models. To the best of our knowledge, we are the first to propose this analysis and categorization for VLN. Code is available at \url{https://github.com/aimagelab/DynamicConv-agent}.
\end{abstract}
%-------------------------------------------------------------------------
\vspace{-.25cm}
\section{Introduction}
\label{sec:intro}
\vspace{-.05cm}
Imagine finding yourself in a large conference hall, with an assistant giving you instructions on how to reach the room for your talk. You are likely to hear something like: \textit{turn right at the end of the corridor, head upstairs and reach the third floor: your room is immediately on the left}. Succeeding in the task of finding your target location is rather nontrivial because of the length of the instruction and its sequential nature: the flow of actions must be coordinated with a series of visual examinations -- like recognizing the end of the corridor or the floor number. Furthermore, navigation complexity dramatically increases if the environment is unknown, and no prior knowledge, such as a map, is available.

%\vspace{0.1cm}
Vision-and-Language Navigation (VLN)~\cite{anderson2018vision} is a challenging task that demands an embodied agent to reach a target location by navigating unseen environments, with a natural language instruction as its only clue. Similarly to the previous example, the agent must assess different sub-tasks to succeed. First, a fine-grained comprehension of the given instruction is needed. Then, the agent must be able to map parts of the description into the visual perception.
For example, \textit{walking past the piano} requires to find and focus on the piano, rather than considering other objects in the scene. Finally, the agent needs to understand when the navigation has been completed and send a stop signal.
\begin{figure}[t!]
    \includegraphics[width=\linewidth]{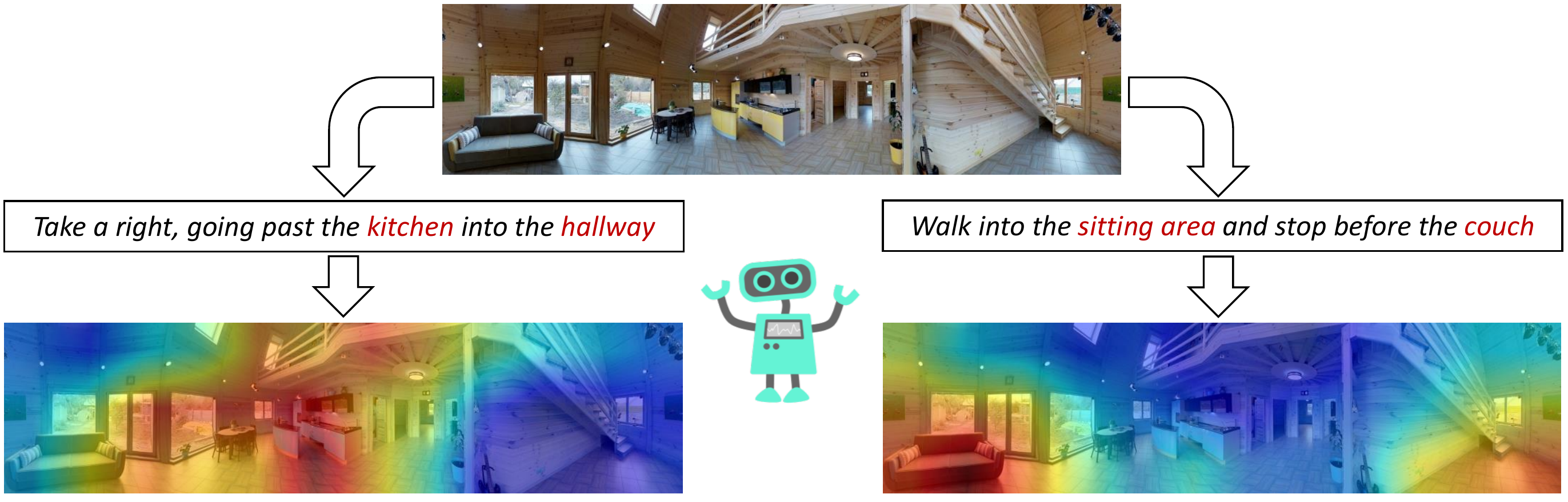}
    \caption{Given a fixed visual observation, dynamic convolutional filters can extract a subset of specific features depending on the leading instruction. In this example, the agent focuses on two different parts of the same environment (best viewed in color).}
    \label{fig:Figure01}
    \vspace{-.5cm}
\end{figure}

%\vspace{0.1cm}
VLN has been first proposed by Anderson \etal~\cite{anderson2018vision}, with the aim of connecting the research efforts on vision-and-language understanding~\cite{vinyals2015show, xu2015show, Anderson_2018, VQA, balanced_vqa_v2, visdial, visdial_rl} with the raising area of embodied AI~\cite{das2018embodied, das2018neural, xia2018gibson, anderson2018evaluation}. This is particularly challenging, as embodied agents must deal with a series of issues that do not belong to traditional vision and language tasks~\cite{anderson2018evaluation}, like contextual decision-making and planning.
Recent works on VLN~\cite{fried2018speaker, ma2019self, ma2019regretful, tan2019learning} integrate the agent with a simplified action space in which it ``\textit{only needs to make high-level decisions as to which navigable direction to go next}''~\cite{fried2018speaker}. In this scenario, the agent does not need to infer the sequence of actions to progress in the environment (\textit{e.g.}, turn right 30 degrees, then move forward) but it exploits a navigation graph to teleport itself to an adjacent location.
The adoption of this high-level action space allowed for a significant boost in success rates, while partly depriving the task of its embodied nature, and leaving space for little more than pure visual and language understanding. We claim that this type of approach is inconvenient, as it strongly relies on prior knowledge on the environment. Depending on information such as the position and the availability of navigable directions, it reduces the task to a pure graph navigation. Moreover, it ignores the active role of the agent, as it only perceives the surrounding scene and selects the next target viewpoint from a limited set. We claim instead that the agent should be the principal component of embodied VLN~\cite{anderson2018evaluation}. Consequently, the output space should match with the low-level set of movements that the agent can perform.
%This explain the great performance improvements of techniques of tree search (such as beam search) that allow to easily make an exhaustive exploration of the whole graph.

%\vspace{0.1cm}
In this paper, we propose a novel architecture for embodied VLN which employs dynamic convolutional filters~\cite{li2017tracking} to identify the next target direction, without getting any information about the navigable viewpoints from the simulator. Convolutional filters are produced via an attention mechanism which follows the given instruction, and are in turn used to attend relevant directions of the scene towards which the agent should move. We then rely on a policy network to predict the sequence of low-level actions. 
%For all of this reasons, we compare our results to Anderson \etal~\cite{anderson2018vision} and Wang \etal~\cite{wang2018look}.

%\vspace{0.1cm}
Dynamic convolutional filters, proposed by Li \etal~\cite{li2017tracking}, were first conceived to identify and track people by a natural language specification. They were then successfully employed in other computer vision tasks, such as actor and action video segmentation from a sentence~\cite{gavrilyuk2018actor}. Nonetheless, these works considered mainly short descriptions, while we deal with complex sentences and long-term dependencies. We generate dynamic filters according to the given instruction, to extract only the relevant information from the visual context. In this way, the same observation can lead to different feature maps, depending on the part of the instruction that the agents must complete (Fig.~\ref{fig:Figure01}).

%Results show that dynamic convolutional filters conditioned on the navigation instruction can improve the performance over the baseline and even outperform state-of-the-art models.
%\vspace{0.1cm}
The proposed method is competitive with prior work that performs high-level navigation exploiting information about the reachable viewpoints (\ie the navigation graph). Additionally, our approach is fully compliant with recent recommendations for embodied navigation~\cite{anderson2018evaluation}.
When compared with models that are compliant with the VLN setup, we overcome the current state of the art by a significant margin.

%\vspace{0.1cm}
\noindent To sum up, our contributions are as follows:
\vspace{-.2cm}
\begin{itemize}
\item We propose a new encoder-decoder architecture for embodied VLN, which for the first time employs dynamic convolutional filters to attend relevant regions of the visual scene and control the actions of the agent.
\vspace{-.3cm}
\item We show, through extensive experimental evaluations, that in a mutable environment with shifting goals dynamic convolutional filters can provide better performance than traditional convolutional filters. Results show that our proposed architecture overcomes the state of the art on the embodied VLN task.
\vspace{-.3cm}
\item As a complementary contribution, we categorize previous work on VLN basing on their level of abstraction and generalizability. We distinguish a group of works that strongly relies on the simulating platform and on the navigation graph, 
we call them \emph{high-level actions} models. A second group, named \emph{low-level actions} models, includes methods that are more agnostic on the underlying implementation and that directly predicts agent actions. With this categorization, we hope to encourage further research to consider low-level and high-level action spaces as distinct fields of application when dealing with VLN.
\end{itemize}

%-------------------------------------------------------------------------
\vspace{-.4cm}
\section{Method}
\label{sec:method}
\vspace{-.2cm}
\begin{figure}[t!]
  \includegraphics[width=\linewidth]{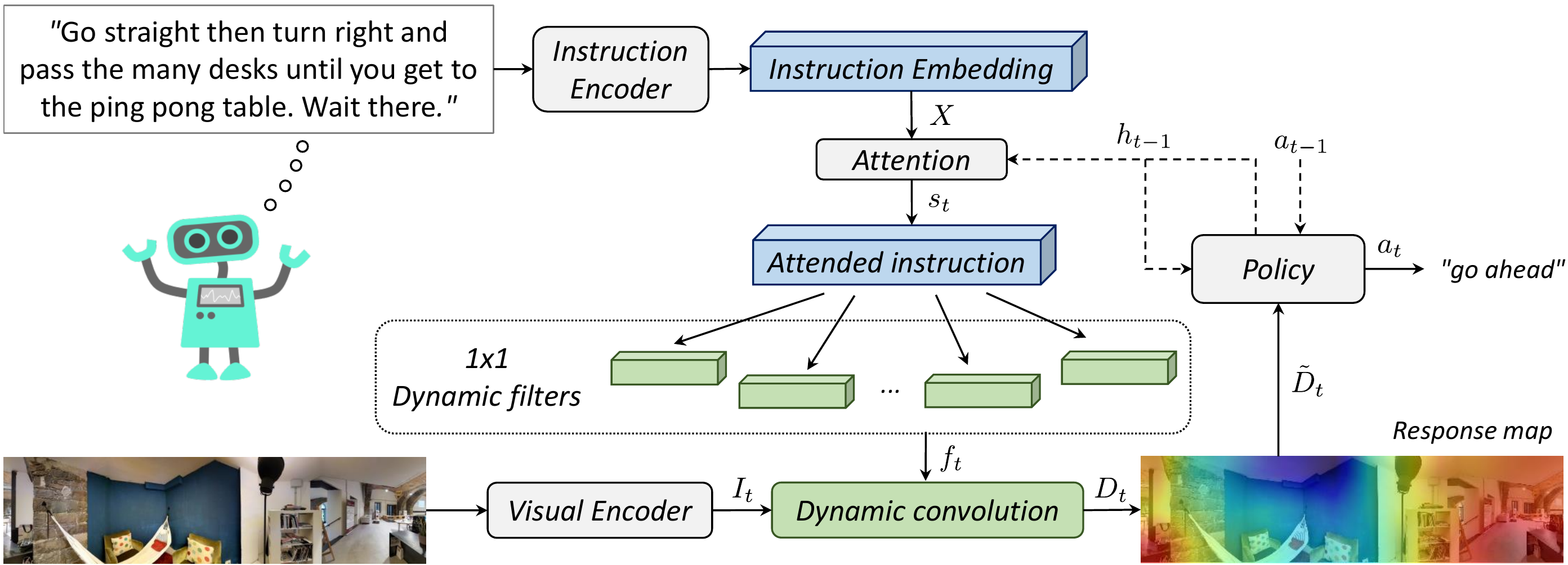}
  \caption{Schema of the proposed architecture for VLN. The input instruction is encoded via an attentive mechanism. We then generate dynamic convolutional filters that let the agent attend relevant regions of the input scene. The resulting visual information is used to feed a policy network controlling the movements of the embodied agent in a low-level action space.}
  \label{fig:model}
  \vspace{-.5cm}
\end{figure}
We propose an encoder-decoder architecture for vision-and-language navigation. Our work employs dynamic convolutional filters conditioned on the current instruction to extract the relevant piece of information from the visual perception, which is in turn used to feed a policy network which controls the actions performed by the agent. The output of our model is a probability distribution over a low-level action space $\bm{A} = \{a_i\}_{i=1}^6$, which comprises the following actions: \textit{turn 30\textdegree~left}, \textit{turn 30\textdegree~right}, \textit{raise elevation}, \textit{lower elevation}, \textit{go ahead}, \textit{<end episode>}. The output probability distribution at a given step, $p_t = P(a_t | X, V_t, h_{t-1})$, depends on a natural language instruction $X$, the current visual observation $V_t$, and on the policy hidden state at time step $t-1$.  Our architecture is depicted in Fig.~\ref{fig:model} and detailed next.

\vspace{-.13cm}
\subsection{Encoder}
\label{subsec:encoder}
To represent the two inputs of the architecture, \ie the instruction and the visual input at time $t$, we devise an instruction and a visual encoder. The instruction encoder provides a representation of the navigation instructions that is employed to guide the whole navigation episode. On the other hand, the visual encoding module operates at every single step, building a representation of the current observation which depends on the agent position.
%\vspace{0.5cm}

%\pagebreak
\textbf{Instruction Encoding.}
The given natural language instruction is split into single words via tokenization, and stop words are filtered out to obtain a shorter description. Differently from previous works that train word embeddings from scratch, we rely on word embeddings obtained from a large corpus of documents. Beside providing semantic information which could not be learned purely from VLN instructions, this also let us handle words that are not present in the training set (see Sec. \ref{subsec:ablation} for a discussion).
Given an instruction with length $N$, we denote its embeddings sequence as $L = \left(l_1, l_2, ..., l_N\right)$, where $l_i$ indicates the embedding for the $i$-th word.
%In our implementation, each word embedding is 300-dimensional. Hence, the full instruction representation has shape [N, 300], where N is the number of words in the filtered sentence. 
Then, we adopt a Long Short-Term Memory (LSTM) network to provide a timewise contextual representation of the instruction:
\begin{equation}
    X = \left(x_1, x_2, ..., x_N\right) = \text{LSTM}(L),
\end{equation}
where each $x_i$ denotes the hidden state of the LSTM at time $i$, thus leading to a final representation with shape $(N, d)$, where $d$ is the size of the LSTM hidden state.

\textbf{Visual Features Encoding.} 
As visual input, we employ the panoramic 360\textdegree~view of the agent, and discretize the resulting equirectangular image in a $12 \times 3$ grid, consisting of three different elevation levels and 30\textdegree~heading shift from each other. Each location of the grid is then encoded via the 2048-dimensional features extracted from a ResNet-152~\cite{he2016deep} pre-trained on ImageNet~\cite{deng2009imagenet}. 
We also append to each cell vector a set of coordinates relative to the current agent heading and elevation:
\begin{equation}
    coord_t = \left(\sin{\phi_t}, \cos{\phi_t}, \sin{\theta_t} \right),
\end{equation}
where $\phi_t \in (-\pi, \pi]$ and $\theta_t \in [-\frac{\pi}{2}, \frac{\pi}{2}]$ are the heading and elevation angles \textit{w.r.t.} the agent position.
By adding $coord_t$ to the image feature map, we encode information related to concepts such as \textit{right}, \textit{left}, \textit{above}, \textit{below} into the agent observation.
%We find this layer beneficial, as not all the feature maps extracted by the pre-trained Resnet-152 are equally useful for our navigation task.

\vspace{-.13cm}
\subsection{Decoder}
\label{subsec:decoder}
Given the instruction embedding $X$ for the whole episode, we use an attention mechanism to select the next part of the sentence that the agent has to fulfill. We denote this encoded piece of instruction as $s_t$. We detail our attentive module in the next section.

\textbf{Dynamic Convolutional Filters.}
Dynamic filters are different from traditional, fixed filters typically used in CNN, as they depend on an input rather than being purely learnable parameters. In our case, we can think about them as specialized feature extractors reflecting the semantics of the natural language specification. For example, starting from an instruction like ``head towards the red chair'' our model can learn specific filters to focus on concepts such as \textit{red} and \textit{chair}. In this way, our model can rely on a large ensemble of specialized kernels and apply only the most suitable ones, depending on the current goal. Naturally, this approach is more efficient and flexible than learning a fixed set of filters for all the navigation steps.
We use the representation of the current piece of instruction $s_t$  to generate multiple $1\times 1$ dynamic convolutional kernels, according to the following equation:
\begin{equation}
    f_t = \ell_2[\text{tanh}(W_f s_t + b_f)],
\end{equation}
where $\ell_2[\cdot]$ indicates L2 normalization, and $f_t$ is a tensor of filters reshaped to have the same number of channels as the image feature map.
We then perform the dynamic convolution over the image features $I_t$, thus obtaining a response map for the current timestep as follows:
\begin{equation}
    D_t = f_t * I_t.
\end{equation}
As the aforementioned operation is equivalent to a dot product, we can conceive the dynamic convolution as a specialized form of dot-product attention, in which $I_t$ acts as key and the filters in $f_t$ act as time-varying queries. Following this interpretation, we rescale $D_t$ by $\sqrt{d_{f}}$, where ${d_{f}}$ is the dynamic filter size~\cite{vaswani2017attention} to maintain dot products smaller in magnitude.

\textbf{Action Selection.}
We use the response maps dynamically generated as input for the policy network. We implement it with an LSTM whose hidden state at time step $t$ is employed to obtain the action scores. Formally,
\begin{equation}
    h_t = \text{LSTM}([\Tilde{D}_t, a_{t-1}], h_{t-1}), \;\;\; p_t = \text{softmax}(W_a h_t + b_a),
\end{equation}
where $[\cdot, \cdot]$ indicates concatenation, $a_{t-1}$ is the one-hot encoding of the action performed at the previous timestep, and $\Tilde{D}_t$ is the flattened tensor obtained from $D_t$.
%The atomic actions that our agent can perform are: turning right or left by 30 degrees, changing its elevation (to go upstairs or downstairs), and moving forward by one step.
To select the next action $a_t$, we sample from a multinomial distribution parametrized with the output probability distribution during training, and select $a_t = \argmax p_t$ during the test. In line with previous work, we find out that sampling during the training phase encourages exploration and improves overall performances.

Note that, as previously stated, we do not employ a high-level action space, where the agent selects the next viewpoint in the image feature map, but instead make the agent responsible of learning the sequence of low-level actions needed to perform the navigation. The agent can additionally send a specific stop signal when it considers the goal reached, as suggested by recent standardization attempts~\cite{anderson2018evaluation}.

\vspace{-.13cm}
\subsection{Encoder-Decoder Attention}
\label{subsec:enc-dec_attention}
The navigation instructions are very complex, as they involve not only different actions but also temporal dependencies between them. Moreover, their high average length represents an additional challenge for traditional embedding methods. For these reasons, we enrich our architecture with a mechanism to attend different locations of the sentence representation, as the navigation moves towards the goal.
% Reference to prior work --- added in camera ready
In line with previous work on VLN~\cite{anderson2018vision, fried2018speaker}, we employ an attention mechanism to identify the most relevant parts of the navigation instruction.
We employ the hidden state of our policy LSTM to get the information about our progress in the navigation episode and extract a time-varying query $q_t = W_q h_{t-1} + b_q$. We then project our sentence embedding into a lower dimensional space to obtain key vectors, and perform a scaled dot-product attention~\cite{vaswani2017attention} among them.
\begin{equation}
    \alpha_t = \frac{q_t K^T}{\sqrt{d_{att}}}, \;\text{where} \; K = W_k X + b_k
\end{equation}
% --- added for camera ready
%According to~\cite{vaswani2017attention}, we find beneficial to rescale the dot product by a factor %$\sqrt{d_{att}}$.
%
After a softmax layer, we obtain the current instruction embedding $s_t$ by matrix multiplication between the initial sentence embedding and the softmax scores.
\begin{equation}
    s_t = \text{softmax}(\alpha_{t}) X
\end{equation}
At each timestep of the navigation process $s_t$ is obtained by attending the instruction embedding at different locations. The same vector is in turn used to obtain a time-varying query for attending spatial locations in the visual input.

\vspace{-.13cm}
\subsection{Training}
\label{subsec:training}
Our training sample consists of a batch of navigation instructions and the corresponding ground truth paths coming from  the R2R (\textit{Room-to-Room}) dataset~\cite{anderson2018vision} (described in section \ref{sec:experiments}). The path denotes a list of discretized viewpoints that the agent has to traverse to progress towards the goal.
The agent spawns in the first viewpoint, and its goal is to reach the last viewpoint in the ground truth list. At each step, the simulator is responsible for providing the next ground truth action in the low-level action space that enables the agent to progress. Specifically, the ground truth action is computed by comparing the coordinates of the next target node in the navigation graph with the agent position and orientation. At each time step $t$, we minimize the following objective function:
\begin{equation}
%    L = -\sum_{t}{y_t \log P(a_t)}
    L = -\sum_{t}{y_t \log{p_t}}
\end{equation}
where $p_t$ is the output of our network, and $y_t$ is the ground truth low-level action provided by the simulator at time step $t$.
We train our network with a batch size of 128 and use Adam optimizer~\cite{kingma2015adam} with a learning rate of $10^{-3}$. We adopt early stopping to terminate the training if the mean success rate does not improve for 10 epochs.
%-------------------------------------------------------------------------
\vspace{-.3cm}
\section{Experiments}
\label{sec:experiments}
\vspace{-.15cm}
\subsection{Experimental Settings}
\label{subsec:exp_setup}
For our experiments, we employ the R2R (\textit{Room-to-Room}) dataset~\cite{anderson2018vision}. This challenging benchmark builds upon Matterport3D dataset of spaces~\cite{Matterport3D} and contains $7,189$ different navigation paths in $90$ different scenes. For each route, the dataset provides 3 natural language instructions, for a total of 21,567 instructions with an average length of 29 words.
The R2R dataset is split into 4 partitions: training, validation on seen environments, validation on unseen scenes, and test on unseen environments.

\textbf{Evaluation Metrics.}
We adopt the same evaluation metrics employed by previous work on the R2R dataset: navigation error (NE), oracle success rate (OSR), success rate (SR), and success rate weighted by path length (SPL). NE is the mean distance in meters between the final position and the goal. SR is fraction of episodes terminated within no more than 3 meters from the goal position. OSR is the success rate that the agent would have achieved if it received an oracle stop signal in the closest point to the goal along its navigation. SPL is the success rate weighted by normalized inverse path length and penalizes overlong navigations.

\textbf{Implementation Details.}
For each LSTM, we set the hidden size to 512. Word embeddings are obtained with GloVe~\cite{pennington2014glove}.
In our visual encoder, we apply a bottleneck layer to reduce the dimension of the image feature map to 512. We generate dynamic filters with 512 channels using a linear layer with dropout~\cite{srivastava2014dropout} ($p=0.5$). In our attention module, $q$ and $K$ have 128 channels and we apply a ReLU non-linearity after the linear transformation. For our action selection, we apply dropout with $p=0.5$ to the policy hidden state before feeding it to the linear layer.

\vspace{-.3cm}
\subsection{Ablation Study}
\label{subsec:ablation}
\begin{table}[t!]
\resizebox{\linewidth}{!}{%
\begin{tabular}{lcccccccc}
\hline
   & \multicolumn{4}{c}{Validation-Seen} & \multicolumn{4}{c}{Validation-Unseen} \\
  \textbf{Method} & NE $\downarrow$ & SR $\uparrow$ & OSR $\uparrow$ & SPL $\uparrow$ 
                  & NE $\downarrow$ & SR $\uparrow$ & OSR $\uparrow$ & SPL $\uparrow$ \\
\hline
Random agent       
& 9.45  & 15.9  & 21.4 & -  & 9.23 & 16.3  & 22.0 & - \\
\hline
Baseline w/ traditional convolution~\cite{anderson2018vision} &
6.01 & 38.6 & 52.9 & - & 7.81 & 21.8 & 28.4 & - \\
Ours w/o encoder-decoder attention & 
5.86 & 41.3 & 51.2 & 36.3 & 7.72 & 22.0 & 29.3 & 19.3 \\
Ours w/o pre-trained embedding &
5.62 & 42.0 & 54.0 & 36.3 & 7.32 & 25.8 & 33.3 & 22.1 \\
\hline
Ours w/ dynamic filters &
\textbf{4.68} & \textbf{53.1} & \textbf{66.1} & \textbf{46.0} &
\textbf{6.65}   & \textbf{31.6}   & \textbf{43.6}   & \textbf{26.8} \\
\hline
\end{tabular}
}\vspace{0.1cm}
\caption{Ablation study for our architecture on the validation sets of R2R. The full model works better than when attention is removed or when conventional filters are used.% The best results are obtained using four different dynamic filters and adopting a pre-trained word embedding to encode the instruction.
}
\label{table:ablation}
\vspace{-.25cm}
\end{table}
\begin{figure}[t!]
    \begin{minipage}[t!]{0.5\textwidth}
        \centering
        \includegraphics[width=0.7\textwidth]{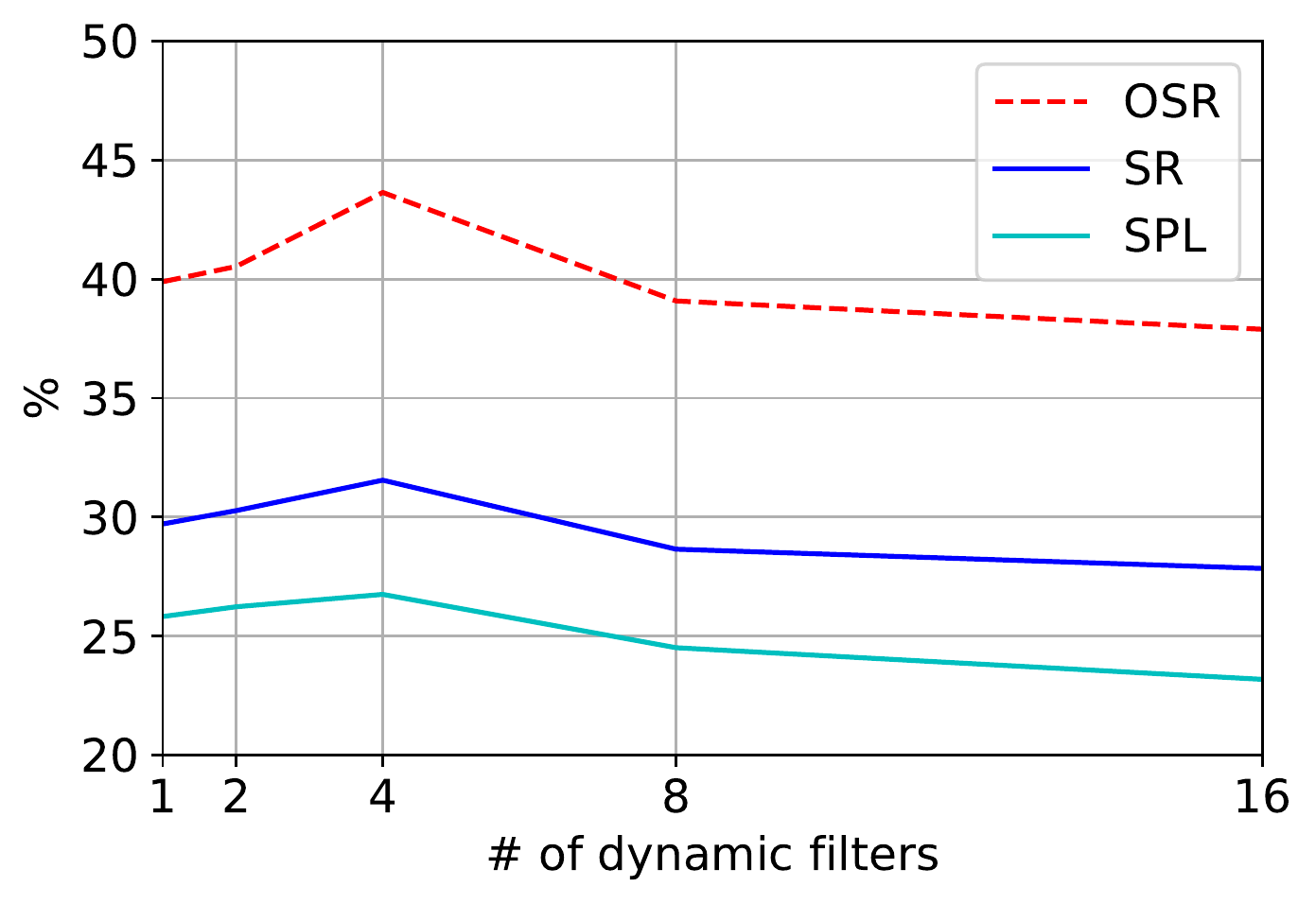}
    \end{minipage}
    \begin{minipage}[t!]{0.45\textwidth}
        \resizebox{!}{0.23\textwidth}{
        \begin{tabular}{c|cccc}
        \hline
         \textbf{\# of} & \multicolumn{4}{c}{\textbf{Validation-Unseen}} \\
         \textbf{filters}  & NE $\downarrow$ & SR $\uparrow$ & OSR $\uparrow$ & SPL $\uparrow$  \\ \hline
         1          & 6.79   & 29.7   & 39.9   & 25.8 \\
         2          & 6.77   & 30.3   & 40.5   & 26.2 \\
         4          & \textbf{6.65}   & \textbf{31.6}   & \textbf{43.6}   & \textbf{26.8} \\
         8          & 7.19   & 28.7   & 39.1   & 24.5 \\
         16         & 7.03   & 27.8   & 37.9   & 23.2 \\
         \hline
\end{tabular}
        }
        \vspace{0.35cm}
    \end{minipage}
    \caption{Comparison with different numbers of dynamic filters on the validation-unseen set of R2R. The best results for all the metrics are obtained using four different dynamic filters.}
    \label{fig:exp}
    \vspace{-.4cm}
\end{figure}
In our ablation study, we test the influence of our implementation choices on VLN. As a first step, we discuss the impact of dynamic convolution by comparing our model with a similar \textit{seq2seq} architecture that employs fixed convolutional filters.
We then detail the importance of using an attention mechanism to extract the current piece of instruction to be fulfilled.
Finally, we compare the results obtained using a pre-trained word embedding instead of learning the word representation from scratch. Results are reported in Table~\ref{table:ablation}.

\textbf{Static Filters Vs. Dynamic Convolution.}
As results show, dynamic convolutional filters surpass traditional fixed filters for VLN. This because they can easily adapt to new instructions and reflect the variability of the task. When compared to a baseline model that employs traditional convolution~\cite{anderson2018vision}, our method performs $14.5\%$ and $9.8\%$ better, in terms of success rate, on the val-seen and val-unseen splits respectively.

\textbf{Fixed Instruction Representation Vs. Attention.}
The navigation instructions are very complex and rich. When removing the attention module from our architecture, we keep the last hidden state $h_N$ as instruction representation for the whole episode. Even with this limitation, dynamic filters achieve better results than static convolution, as the success rate is higher for both of the validation splits. However, our attention module further increases the success rate by $11.8\%$ and $9.6\%$.

\textbf{Word Embedding from Scratch Vs. Pre-trained Embedding.}
Learning a meaningful word embedding is nontrivial and requires a large corpus of natural language descriptions. For this reason, we adopt a pre-trained word embedding to encode single words in our instructions. We then run the same model while trying to learn the word embedding from scratch. We discover that a pre-trained word embedding significantly eases VLN. Our model with GloVe~\cite{pennington2014glove} obtains $11.1\%$ and $5.8\%$ more on the val-seen and val-unseen splits respectively, in terms of success rate.

\subsection{Multi-headed Dynamic Convolution}
\label{subsec:exp3}
\vspace{-.05cm}In this experiment, we test the impact of using a different number of dynamically-generated filters. We test our architecture when using 1, 2, 4, 8, and 16 dynamic filters. We find out that the best setup corresponds to the use of 4 different convolutional filters. Results in Fig.~\ref{fig:exp} show that the success rate and the SPL increase linearly with the number of dynamic kernels for a small number of filters, reaching a maximum at 4. The metrics then decrease when adding new parameters to the network. This suggests that a low number of dynamic filters can represent a wide variety of natural language specifications. However, as the number of dynamic filters increase, the representation provided by the convolution becomes less efficient.
%We hypothesize that, when using a large number of filters, dynamic convolution is equivalent in performance to static convolution with fixed kernels.

\vspace{-.25cm}
\subsection{Comparison with the State-of-the-art}
\label{subsec:comparison}
\begin{table}[t!]
\resizebox{\linewidth}{!}{%
\begin{tabular}{ccccccccccccc}
\hline
                 & \multicolumn{4}{c}{Validation-Seen} & \multicolumn{4}{c}{Validation-Unseen} & \multicolumn{4}{c}{Test (Unseen)} \\
\textbf{Low-level Actions Methods} & NE $\downarrow$ & SR $\uparrow$ & OSR $\uparrow$ & SPL $\uparrow$    &
NE $\downarrow$ & SR $\uparrow$ & OSR $\uparrow$ & SPL $\uparrow$  &
NE $\downarrow$ & SR $\uparrow$ & OSR $\uparrow$ & SPL $\uparrow$ \\
\hline
Random           
& 9.45 & 0.16 & 0.21 & -     & 9.23 & 0.16 & 0.22 & -    & 9.77 & 0.13 & 0.18 & 0.12   \\
Student-forcing~\cite{anderson2018vision}
& 6.01 & 0.39 & 0.53 & -     & 7.81 & 0.22 & 0.28 & -    & 7.85 & 0.20 & 0.27 & 0.18   \\
RPA~\cite{wang2018look}
& 5.56 & 0.43 & 0.53 & -     & 7.65 & 0.25 & 0.32 & -    & 7.53 & 0.25 & 0.33 & 0.23   \\
\hline
Ours 
& 4.68 & 0.53 & 0.66 & 0.46     & 6.65 & 0.32 & 0.44 & 0.27      & 7.14 & 0.31 & 0.42 & 0.27 \\
Ours w/ data augmentation
& \textbf{3.96} & \textbf{0.58} &\textbf{ 0.73} & \textbf{0.51}     
& \textbf{6.52} & \textbf{0.34} & \textbf{0.43} & \textbf{0.29}
& \textbf{6.55} & \textbf{0.35} & \textbf{0.45} & \textbf{0.31} \\
\hline
\\
\textbf{High-level Actions Methods} & NE $\downarrow$ & SR $\uparrow$ & OSR $\uparrow$ & SPL $\uparrow$    &
NE $\downarrow$ & SR $\uparrow$ & OSR $\uparrow$ & SPL $\uparrow$  &
NE $\downarrow$ & SR $\uparrow$ & OSR $\uparrow$ & SPL $\uparrow$ \\
\hline
Speaker-Follower~\cite{fried2018speaker}
& 3.36 & 0.66 & 0.74 & -     & 6.62 & 0.36 & 0.45 & -    & 6.62 & 0.35 & 0.44 & 0.28   \\
%RCM \cite{wang2018reinforced}
%& 3.37    & 0.67    & 0.77   & -      & 5.88    & 0.43    & 0.52    & -       & 6.01   & 0.43   & 0.51   & 0.35   %\\
Self-Monitoring~\cite{ma2019self}
& \textbf{3.22}    & 0.67    & \textbf{0.78}   & 0.58   & 5.52    & 0.45    & 0.56    & 0.32    & 5.99   & 0.43   & 0.55   & 0.32   \\
Regretful~\cite{ma2019regretful}
& 3.23 & \textbf{0.69} & 0.77 & \textbf{0.63}  &
\textbf{5.32} & \textbf{0.50} & \textbf{0.59} & \textbf{0.41} &
\textbf{5.69} & \textbf{0.48} & \textbf{0.56} & \textbf{0.40}   \\
\hline
\end{tabular}}\vspace{0.1cm}
\caption{Comparison with state-of-art methods for VLN. For compliance with the evaluation server, we report success rates as fractions. The results for high-level models comprehend data augmentation with synthetic data provided by~\cite{fried2018speaker}, as in our final setup. Our method outperforms comparable models by a large margin, while being competitive with or even better than some high-level actions architectures.}
\label{table:sota_comparison}
\vspace{-.4cm}
\end{table}
\vspace{-.05cm}Finally, we compare our architecture with the state-of-the-art methods for VLN. Results are reported in Table~\ref{table:sota_comparison}.
We distinguish two main categories of models, depending on their output space: the first, to which our approach belongs, predicts the next atomic action (\eg \textit{turn right}, \textit{go ahead}). We call architectures in this category \textit{low-level actions methods}. The second, instead, searches in the visual space to match the current instruction with the most suitable navigable viewpoint. In these models, atomic actions are not considered, as the agent displacements are done with a teleport system, using the next viewpoint identifier as target destination. Hence, we refer to these works as \textit{high-level actions methods}.
While the latter achieve better results, they make strong assumptions on the underlying simulating platform and on the navigation graph.
Our method, exploiting dynamic convolutional filters and predicting atomic actions, outperforms comparable architectures and achieves state of the art results for \textit{low-level actions} VLN. Our final implementation takes advantage of the synthetic data provided by Fried \etal~\cite{fried2018speaker} and overcomes comparable methods~\cite{anderson2018vision, wang2018look} by $15\%$ and $10\%$ success rate points on the R2R test set.
Additionally, we note that our method is competitive with some \textit{high-level actions} models, especially in terms of SPL. When considering the test set, we notice in fact that our model outperforms Speaker-Follower~\cite{fried2018speaker} by $3\%$, while performing only $1\%$ worse than~\cite{ma2019self}.

\textbf{Low-level Action Space or High-level Navigation Space?}
While previous work on VLN never considered this important difference, we claim that it is imperative to categorize navigation architectures depending on their output space.
In our opinion, ignoring this aspect would lead to inappropriate comparisons and wrong conclusions.
Considering the results in Table~\ref{table:sota_comparison}, we separate the two classes of work and highlight the best results for each category. Please note that the random baseline was initially provided by \cite{anderson2018vision} and belongs to \textit{low-level actions} architectures (a random \textit{high-level actions} agent was never provided by previous work).
We immediately notice that, with this new categorization, intra-class results have less variance and are much more aligned to each other.
We believe that future work on VLN should consider this new taxonomy in order to provide meaningful and fair comparisons.
\subsection{Qualitative Results}
\label{subsec:qualitative}
\begin{figure}[h!]
    \centering
    \begin{tabular}{ccccc}
% Actions legend
    \small{\textbf{Legend: }} &
    \includegraphics[width=0.025\textwidth]{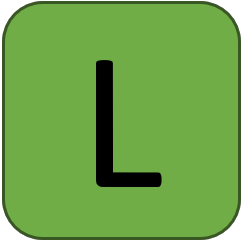} \small{\textit{left}} &
    \includegraphics[width=0.025\textwidth]{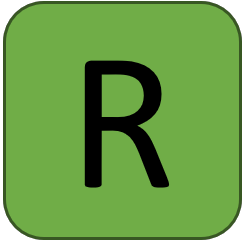} \small{\textit{right}} & 
    \includegraphics[width=0.025\textwidth]{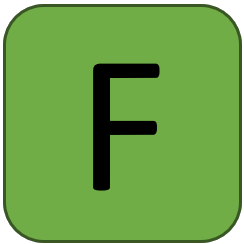} \small{\textit{forward}} &
    \includegraphics[width=0.025\textwidth]{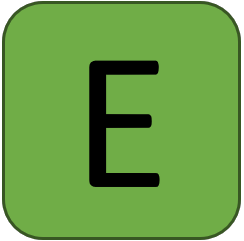} \small{\textit{end episode}} \\\vspace{-.25cm}\\
    \end{tabular}
    \begin{tabular}{cccc}
% Qualitative #1
    \includegraphics[width=0.2\textwidth]{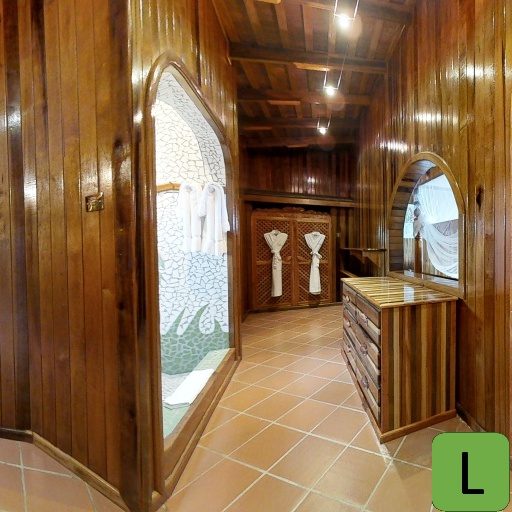} &
    \includegraphics[width=0.2\textwidth]{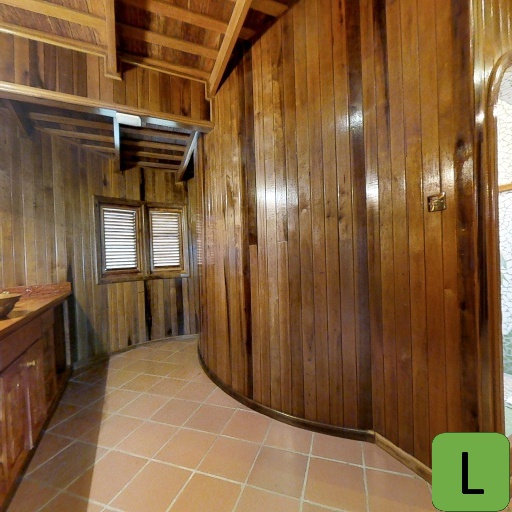} &
    \includegraphics[width=0.2\textwidth]{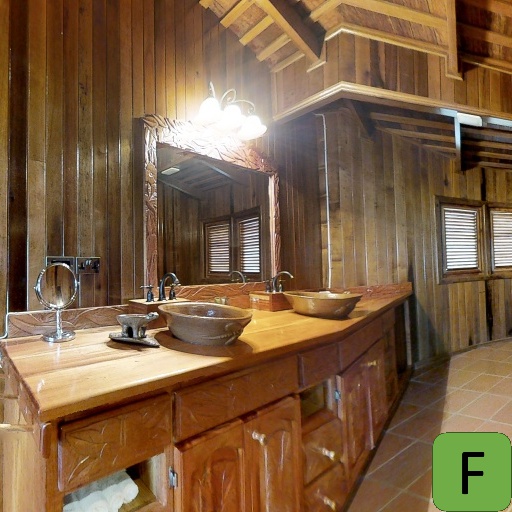} &
    \includegraphics[width=0.2\textwidth]{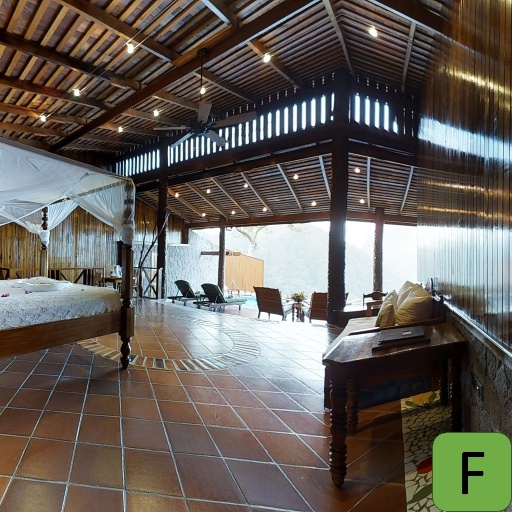} \\
    \includegraphics[width=0.2\textwidth]{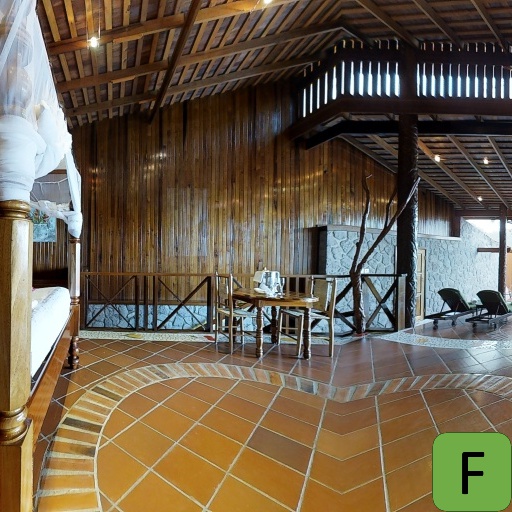} &
    \includegraphics[width=0.2\textwidth]{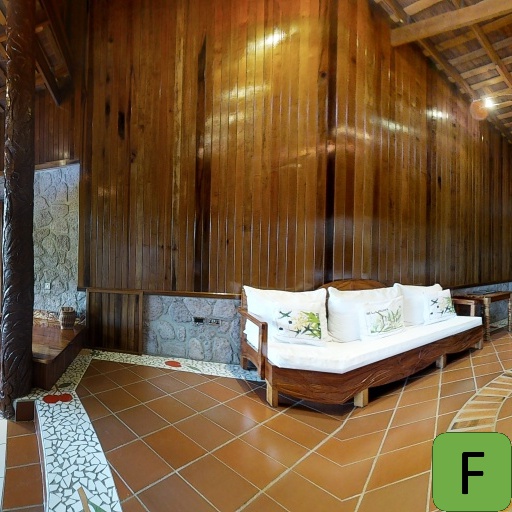} &
    \includegraphics[width=0.2\textwidth]{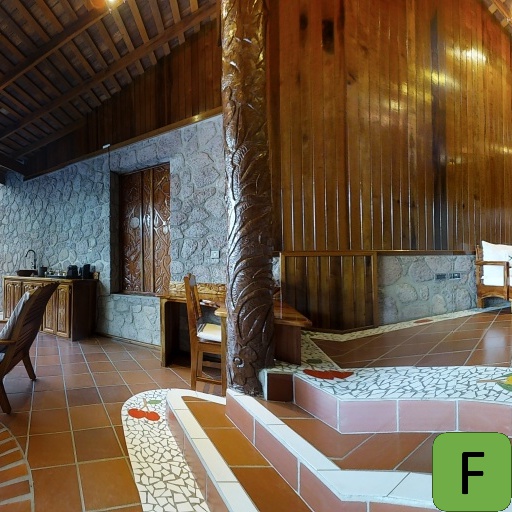} &
    \includegraphics[width=0.2\textwidth]{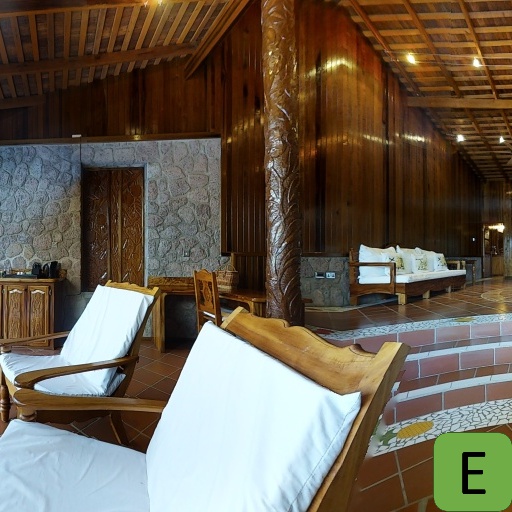} \\
    \multicolumn{4}{c}{\small{\textbf{Instruction:} \textit{From bathroom, enter bedroom and walk straight}}}\\
    \multicolumn{4}{c}{\small{\textit{across down two steps, wait at loungers.}}}   \\\vspace{-.25cm}\\ 
%
% Qualitative #2
    \includegraphics[width=0.2\textwidth]{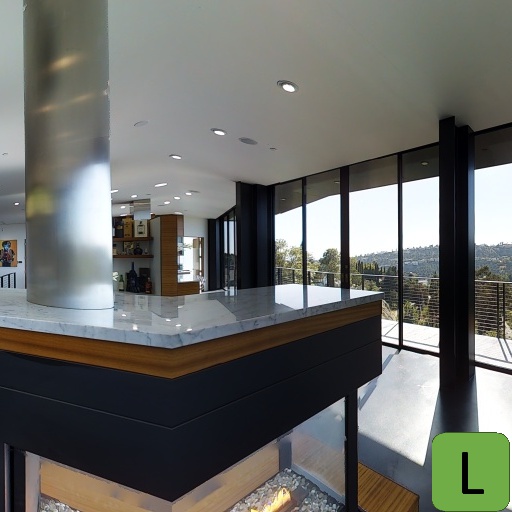} &
    \includegraphics[width=0.2\textwidth]{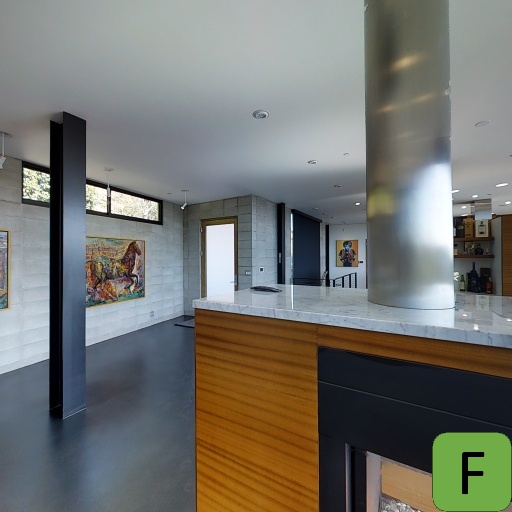} &
    \includegraphics[width=0.2\textwidth]{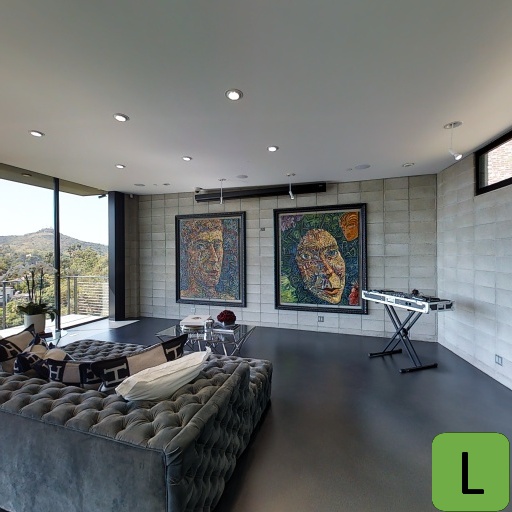} &
    \includegraphics[width=0.2\textwidth]{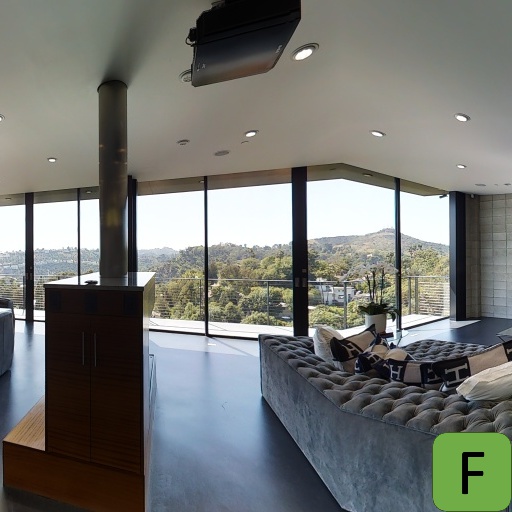} \\
    \includegraphics[width=0.2\textwidth]{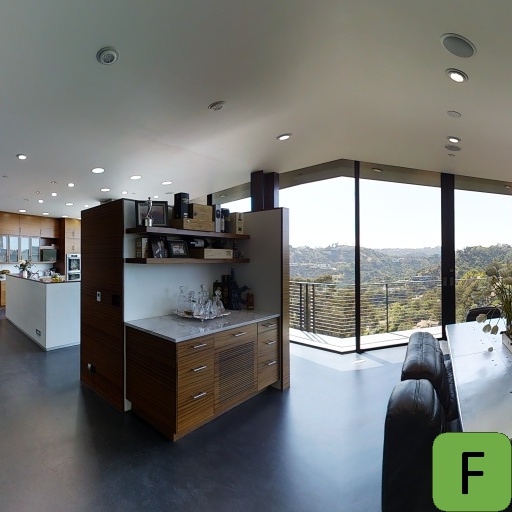} &
    \includegraphics[width=0.2\textwidth]{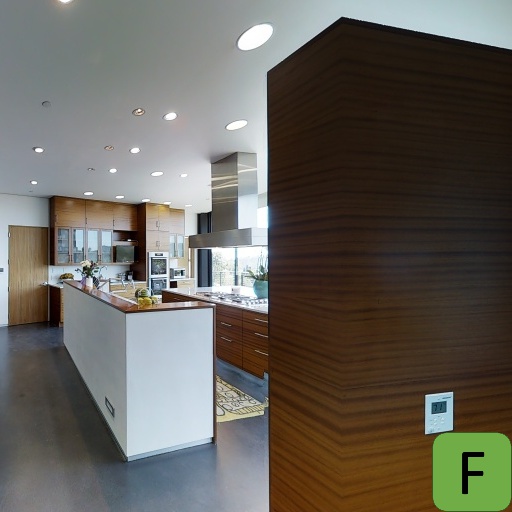} &
    \includegraphics[width=0.2\textwidth]{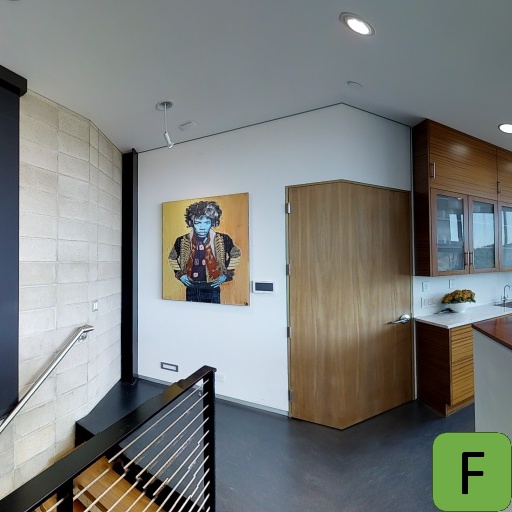} &
    \includegraphics[width=0.2\textwidth]{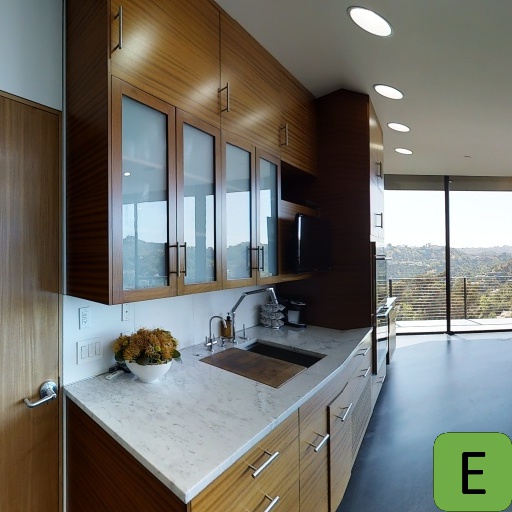} \\
    \multicolumn{4}{c}{\small{\textbf{Instruction:} \textit{Walk past the fireplace and to the left.}}}\\
    \multicolumn{4}{c}{\small{\textit{Stop in the entryway of the kitchen.}}}   \\
    \end{tabular}
    \vspace{0.2cm}
    \caption{Qualitative results from the R2R validation set. Each episode is detailed by eight pictures, representing the current position of the agent and containing the next predicted action (from left to right, top to bottom). To make the visualization more readable, we do not display the 360\textdegree~panoramic images.}
    \label{fig:qualitative_1}
    \vspace{-.4cm}
\end{figure}

Fig.~\ref{fig:qualitative_1} shows two navigation episodes from the R2R validation set. We display the predicted action in a green box on the bottom-right corner of each image. Both examples are successful.
\vspace{-.3cm}
%-------------------------------------------------------------------------
\section{Conclusion}
\label{sec:conclusion}
\vspace{-.25cm}
In this paper, we propose dynamic convolution for embodied Vision-and-Language Navigation. Instead of relying on a high-level action space, where the agent is teleported from one viewpoint to the other, we predict a series of action in an agent friendly action space.
Basing on this substantial difference, we propose a new categorization based on the model output space. We then separate previous VLN architectures into \textit{low-level actions} and \textit{high-level actions} methods. We claim that comparisons made considering this new taxonomy are more fair and reasonable than previous analysis.
%discuss previous research on VLN and recognize two main directions of work.
%The first treats the navigation problem as a graph exploration task, relying on the structure of the underlying simulator, and outputs a series of target viewpoints. The second provides instead a sequence of atomic actions in a low-level action space. Following this observation, we propose a new categorization based on the model's output space. We then separate previous VLN architectures into \textit{low-level actions} and \textit{high-level actions} methods. We claim that comparison made considering this new taxonomy are more fair and reasonable than previous analysis.
Our method with dynamic convolutional filters achieves state-of-the-art results for the \textit{low-level actions} category, and it is competitive with \textit{high-level actions} architectures that rely on much more information and have a higher level of abstraction during the navigation episode.
% Future works -- added in camera ready version 
We hope this work encourages further research on low-level VLN, and in general we consider this a step towards the use of more realistic action spaces for this task.
While our experiments show promising results in this setting, much work remains to inspect the possible connections between low-level and high-level Vision-and-Language Navigation.

\hyphenation{Ri-spar-mio}

\vspace{0.3cm}
\noindent\textbf{Acknowledgements:} This work was partially supported by the \textit{Fondazione Cassa di Risparmio di Modena} project ``\textit{AI for Digital Humanities}'' (Prot. n. 505.18.8b del 18/10/2018 - Pratica Sime n. 2018.0390). We also want to thank the anonymous reviewers for their insightful remarks and their constructive criticism.

%-------------------------------------------------------------------------
%\bibliography{egbib}
\bibliography{references}
\end{document}